\documentclass[pmlr,twocolumn,10pt]{jmlr} 

\usepackage{booktabs}
\usepackage{siunitx}
\usepackage{caption}
\usepackage{subcaption}
\usepackage{dsfont}

\SetCommentSty{mycommfont}

\usepackage[switch]{lineno}

\DeclareMathOperator\erf{erf}

\theorembodyfont{\upshape}
\theoremheaderfont{\scshape}
\theorempostheader{:}
\theoremsep{\newline}

\title{Learning Social Fairness Preferences from Non-Expert Stakeholder Opinions in Kidney Placement}

\author{%
\Name{Mukund Telukunta} \Email{mt3qb@mst.edu}\\
\addr Missouri University of Science and Technology
\AND
\Name{Sukruth Rao} \Email{raosukru@msu.edu}\\
\addr Michigan State University
\AND
\Name{Gabriella Stickney} \Email{stickn21@msu.edu}\\
\addr Michigan State University
\AND
\Name{Venkata Sriram Siddardh Nadendla} \Email{nadendla@mst.edu}\\
\addr Missouri University of Science and Technology
\AND
\Name{Casey Canfield} \Email{canfieldci@mst.edu}\\
\addr Missouri University of Science and Technology
}

\begin{document}

\maketitle
\thispagestyle{plain}

\begin{abstract}
Modern kidney placement incorporates several intelligent recommendation systems which exhibit social discrimination due to biases inherited from training data. Although initial attempts were made in the literature to study algorithmic fairness in kidney placement, these methods replace true outcomes with surgeons' decisions due to the long delays involved in recording such outcomes reliably. However, the replacement of true outcomes with surgeons' decisions disregards expert stakeholders' biases as well as social opinions of other stakeholders who do not possess medical expertise. This paper alleviates the latter concern and designs a novel fairness feedback survey to evaluate an acceptance rate predictor (ARP) that predicts a kidney's acceptance rate in a given kidney-match pair. The survey is launched on Prolific, a crowdsourcing platform, and public opinions are collected from 85 anonymous crowd participants. A novel social fairness preference learning algorithm is proposed based on minimizing social feedback regret computed using a novel logit-based fairness feedback model. The proposed model and learning algorithm are both validated using simulation experiments as well as Prolific data. Public preferences towards group fairness notions in the context of kidney placement have been estimated and discussed in detail. The specific ARP tested in the Prolific survey has been deemed fair by the participants.
\end{abstract}


\paragraph*{Data and Code Availability}
This paper uses the kidney matching dataset (STAR file) requested from the Organ Procurement and Transplant Network (OPTN) to generate the data tuples presented to the survey participants. Given the sensitivity of data used in both simulation experiments as well as survey dataset, both the code and dataset are also not released to the public. However, both code and data can be made available upon request only after obtaining consent from OPTN to avail the STAR file. 

\paragraph*{Institutional Review Board (IRB)}
This research paper has undergone ethical review and approval by the IRB with the approval number 2092366. The informed consent process, including the information provided to participants and the procedures for obtaining their voluntary and informed consent, has been reviewed and approved by the IRB. Participants were assured of the confidentiality and privacy of their data, and all efforts have been made to minimize any potential risks associated with their involvement in the study.

\section{Introduction}
The increasing rate of kidney discard in deceased donors \citep{lentine2023optn} has inspired the adoption of
machine learning (ML) solutions 
to 
identify kidneys with high discard risk \citep{barah2021predicting}, provide analytics on kidney offer acceptance decisions \citep{mcculloh2023experiment}, and offer recommendations to surgeons by predicting the acceptance of a donor kidney \citep{ashiku2022identifying}. However, these models are susceptible to social discrimination, as they are trained using past decisions curated during traditional kidney placement practices. For instance, the inclusion of \emph{race} coefficient in the computation of Kidney Donor Profile Index (KDPI) systematically assigns higher scores to kidneys from Black donors irrespective of whether or not they carry the APOL1 gene (one that results in a guaranteed failure of renal transplantation), thereby contributing to an increase in the overall discard rate \citep{chong2021reconsidering}. At the same time, the \emph{age} attribute in calculating patient's Estimated Post Transplant Survival (EPTS) score allocates high-quality kidneys to younger recipients at the expense of older patients with a potentially greater medical need \citep{eidelson2012kidney}. Therefore, there is an urgent need to quantify the fairness of such ML-based systems using mathematical fairness notions.  

Unfortunately, a significant limitation with state-of-the-art fairness notions (especially group-based notions \citep{mehrabi2021survey}) is their reliance on final outcomes, which are usually observed in hindsight. For example, the death of an organ recipient can only be observed in hindsight, only during a two-year post transplantation monitoring period. The process of recording true outcomes is very challenging due to the need to track organ recipients post surgery over at least 2-5 years. As an alternative, human perception of fairness is proposed based on perceived labels which are collected from expert critics for a quick analysis \citep{srivastava2019mathematical, grgic2018human}. However, such an approach is myopic in nature, as it does not take into account other stakeholders' opinions, which could differ quite significantly from medical experts' opinions.

The stakeholders in kidney placement can be broadly classified into two types: (i) \emph{clinical experts} are those with medical expertise to recommend/authorize kidney offer decisions (e.g. transplant surgeons, organ procurement teams), and (ii) \emph{personal experts} are those who lack technical knowledge but possess the basic understanding through interaction with clinical experts as well as their own peers (e.g. donors/recipients, their friends and family). Although clinical experts evaluate the likelihood of recipient's post-transplant survival based on available medical data, they are seldom available for feedback elicitation. On the contrary, personal experts and public critics are available freely and always express their eagerness to express opinions and fairness preferences. This paper focuses on the learning of social preference across diverse group-fairness notions.

The main contributions of this paper are three-fold. Firstly, this paper investigates the \textbf{\emph{first-of-its-kind non-expert (i.e., public) perception of fairness of ML-based models used in kidney placement}} pipeline. A human-subject \textbf{\emph{survey experiment}} was conducted on Prolific crowdsourcing platform to collect feedback regarding the fairness of a ML-based system from non-expert (public) participants. In contrast to prior efforts, participants are not constrained to any particular fairness perspective, and are free to choose their preferred group fairness notions at will, and assess the fairness of the ML-system for a given sensitive attribute(s). Secondly, a \textbf{\emph{novel logit-based feedback model}} is proposed based on encoded Likert choices and \emph{ambiguous fairness preferences} across group fairness notions. Thirdly, a \textbf{\emph{projected gradient-descent algorithm with an efficient gradient computation}} is designed to minimize social feedback regret. The proposed approach is validated on a wide range of simulation experiments. Finally, the proposed method was adopted to analyze and \textbf{\emph{find public's social preferences recorded in Prolific survey}} dataset.  

The remainder of this paper is organized as follows. Section \ref{Sec: Literature Survey - Human Fairness Perception} presents a brief literature survey on human fairness perception. The Prolific experiment is discussed in Section \ref{Sec: Experiments}, which is then followed by the proposed methodology in Section \ref{sec: Methodology}. Evaluation methodology is presented in detail in Section \ref{sec: Evaluation}, followed by results and their discussion in Section \ref{sec: Results}.

\begin{figure*}[!t]
\centering
\includegraphics[width=\textwidth]{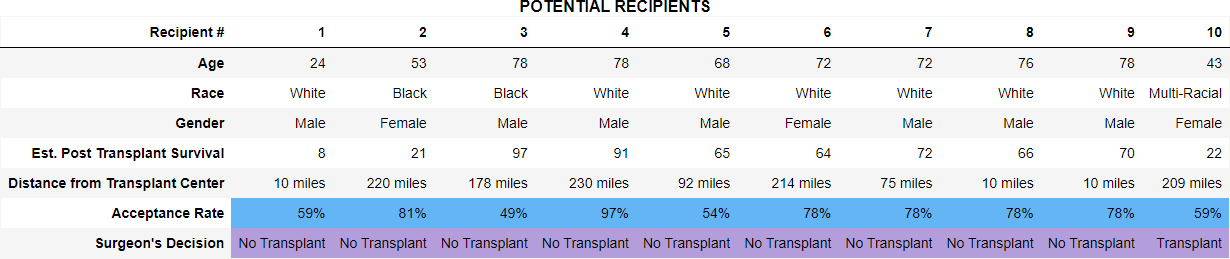}
\caption{An Example of Recipient Characteristics}
\label{Fig: Recipient Characteristics}
\end{figure*}

\begin{figure*}[!t]
\centering
\includegraphics[width=\textwidth]{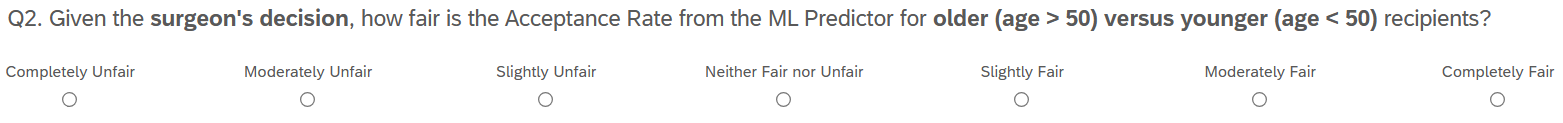}
\includegraphics[width=\textwidth]{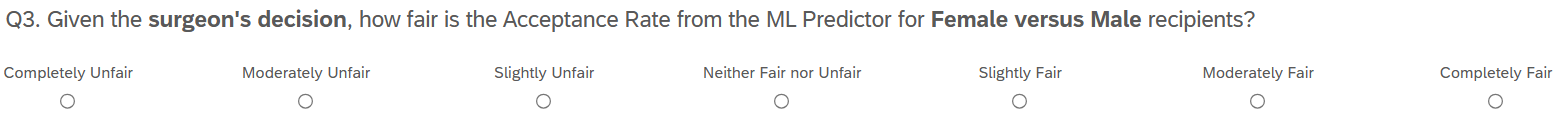}
\includegraphics[width=\textwidth]{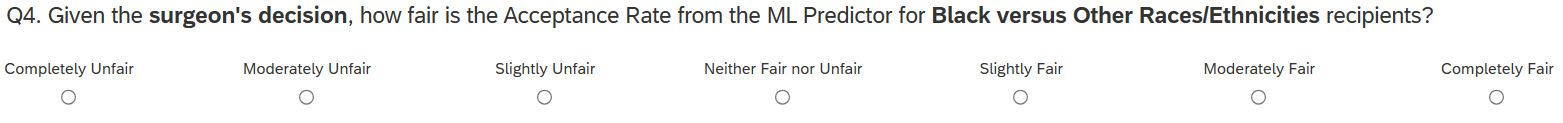}
\caption{Four Questions Presented to the Participants for Each Data Tuple}
\label{Fig: Survey Questions}
\end{figure*}

\section{Human Fairness Perception: A Brief Literature Survey \label{Sec: Literature Survey - Human Fairness Perception}}
In the past, several researchers have attempted to model human perception of fairness. For instance, in an experiment performed by \cite{srivastava2019mathematical}, participants were asked to choose among two different models to identify which notion of fairness (demographic parity or equalized odds) best captures people's perception in the context of both risk assessment and medical applications. Likewise, another team surveyed 502 workers on Amazon's Mturk platform and observed a preference towards \emph{equal opportunity} in \cite{harrison2020empirical}. Work by \cite{grgic2018human} discovered that people's fairness concerns are typically multi-dimensional (relevance, reliability, and volitionality), especially when binary feedback was elicited. A very recent work of \cite{lavanchy2023applicants} conducts four survey experiments to study applicants' perception towards algorithm-driven hiring procedures. Their findings indicate that recruitment processes are deemed less fair compared to human only or AI-assisted human processes, regardless of applicants receiving a positive outcome.

\begin{table}[!t]
\centering
\begin{tabular}{l l}
\toprule
\multicolumn{2}{c}{\textbf{DONOR}}
\\ \midrule
\textbf{Age} & 28
\\
\textbf{Race} & White
\\
\textbf{Gender} & Male
\\
\textbf{Kidney Quality} & 18
\\\bottomrule
\end{tabular}
\caption{An Example of Donor Characteristics}
\label{Tab: DD Characteristics}
\end{table}

\section{Experiment Design \label{Sec: Experiments}}
The objective of the survey experiment is to collect non-expert (i.e. public) feedback regarding the fairness of a state-of-the-art kidney acceptance rate predictor (ARP) \citep{ashiku2022identifying}. This predictor is an analytics tool that predicts kidney acceptance probability based on donor-recipient characteristics (includes both medical features and social demographics) in order to support transplant surgeon decisions regarding deceased donor kidney offers and alleviate kidney discards. The predictor was trained using kidney matching datasets spanning from 2014 to 2018, achieving a testing accuracy of 96\%. 

\subsection{Datasets and Preprocessing}
Public participants are provided with predictions from the ARP for various kidney matching instances spanning 2020 and 2021. These predictions are based on datasets called Standard Transplant Analysis and Research (STAR) files, obtained from the Organ Procurement and Transplant Network (OPTN). The STAR files contain anonymized patient-level data on transplant recipients, donors, and matches dating back to 1987. Each dataset typically includes numerous instances where a deceased donor kidney is matched with thousands of potential recipients. Since presenting such large datasets can overwhelm the participants, the number of potential recipients for each deceased donor was limited to $K = 10$. This subset includes at least one recipient who received the kidney, ensuring a balanced representation of successful and unsuccessful transplant outcomes. The remaining recipients were randomly selected. Additionally, recipients under 17 years old were excluded due to unique challenges in pediatric transplantation \citep{magee2004pediatric}. The preprocessed dataset comprised 13,628 deceased donors from 2021 and 5,023 from 2022. A sample of $M = 10$ deceased donors (7 from 2021 and 3 from 2022) was randomly selected from the preprocessed STAR dataset. The ARP was then applied to this sample to obtain acceptance rates for every potential recipient within each deceased donor kidney. A single donor paired with 10 potential recipients is considered as a \emph{data-tuple}.

\subsection{Survey Questions}
This survey presents data as two distinct tables for each data-tuple. The first table contains information regarding the deceased donor including donor's age, race, gender, and KDPI score. As an illustration, Table \ref{Tab: DD Characteristics} presents the donor characteristics in a data tuple example presented to the survey participant. The second table presents information on ten recipient profiles matched with this donor, which includes each recipient's age, race, gender, EPTS score, distance from the transplant center, prediction from ARP, and the surgeon's decision (transplant or no transplant), as shown in the Figure \ref{Fig: Recipient Characteristics}. Subsequently, the participants were instructed to respond to four distinct questions within each data-tuple. Initially, they were asked to rate the fairness of the ARP using a Likert scale ranging from 1 to 7 (denoted as $s$), where 1 indicates complete unfairness, and 7 denotes complete fairness. Following this, the participants were further prompted to assess the fairness of the ARP in context of (i) older recipients (age $> 50$) versus younger recipients (age $< 50$), (ii) female versus male recipients, and (iii) Black recipients versus recipients from other racial backgrounds (as shown in Figure \ref{Fig: Survey Questions}).

\begin{table}[!t]
\centering
\begin{tabular}{l|r|r}
\hline
\textbf{Demographic Attribute} & \textbf{Prolific} & \textbf{Census}
\\ \hline
\\[-2,5ex]
18-25 & 8\% & 13\%
\\[-0.5ex]
25-40 & 57\% & 26\%
\\[-0.5ex]
40-60 & 29\% & 32\%
\\[-0.5ex]
$>$60 & 6\% & 22\%
\\ \hline
\\[-2.5ex]
White & 60\% & 59\%
\\[-0.5ex]
Black & 19\% & 12\%
\\[-0.5ex]
Asian & 12\% & 5.6\% 
\\[-0.5ex]
Hispanic & 3.4\% & 18\%
\\[-0.5ex]
Other & 5.6\% & 9\%
\\ \hline
\\[-2.5ex]
Male & 49\% & 49.5\%
\\[-0.5ex]
Female & 49\% & 50.5\%
\\[-0.5ex]
Non-binary & 2\% & -
\\ \hline
\\[-2.5ex]
High School or equivalent & 18\% & 26.5\%
\\[-0.5ex]
Bachelor's (4 year) & 40\% & 20\%
\\[-0.5ex]
Associate (2 year) & 15\% & 8.7\%
\\[-0.5ex]
Some college & 12\% & 20\%
\\[-0.5ex]
Master's & 11\% & 13\%
\\[-0.5ex]
\end{tabular}
\caption{Participants demographics compared to the 2021 U.S. Census Data.}
\label{Tab: participant demographics}
\end{table}

\subsection{Participant Demographics}
The survey experiment was deployed on Prolific (IRB Reference Number 2092366) during December 2023. A total of 85 participants were recruited for the study. Among them, $N = 75$ individuals were chosen, with the exclusion of 8 participants experiencing technical difficulties, and an additional 2 participants failing to answer the attention check questions. Table \ref{Tab: participant demographics} summarizes the demographics of the recruited participants. The recruited participants consisted of fewer Hispanics (3.4\%), more Blacks (19\%), more educated (51\%) and more younger (65\%) individuals compared to the 2021 U.S. Census \citep{Census2021ACS}. 

\section{Methodology\label{sec: Methodology}}

\subsection{Fairness Feedback Model \label{Sec: Feedback Model}}
\begin{figure}[!t]
\centering
\includegraphics[width = 0.45\textwidth]{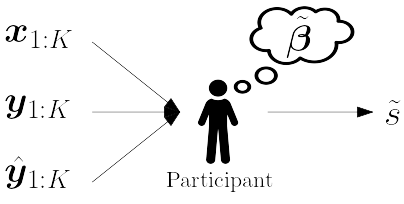}
\caption{Non-Expert Participant's Feedback Model}
\label{Fig: Nonexpert Feedback Model}
\end{figure}

Consider $N$ non-expert participants who evaluate the acceptance rate predictor (ARP) 
from the perspective of group fairness across sensitive demographics. The $n^{th}$ participant investigates the $m^{th}$ representative \emph{data-tuple} $\boldsymbol{d}_m = \{ \boldsymbol{x}_{1:K}^{(m)}, \boldsymbol{y}_{1:K}^{(m)}, \hat{\boldsymbol{y}}_{1:K}^{(m)} \}$ from ARP, which comprises of the donor-recipient attributes $\boldsymbol{x}_{1:K}^{(m)}$, surgeon's decisions $\boldsymbol{y}_{1:K}^{(m)}$ and the ARP's predictions $\hat{\boldsymbol{y}}_{1:K}^{(m)}$ across $K$ donor-recipient pairs. Upon investigation, the $n^{th}$ participant presents a fairness feedback score $s_{n,m} \in \{1, 2, \cdots, 7\}$ to the evaluation platform (as depicted in Figure \ref{Fig: Nonexpert Feedback Model}), where $s_{n,m} = 1$ indicates an unfair ARP and $s_{n,m} = 7$ indicates a fair ARP. 

In this section, the $n^{th}$ participant's fairness feedback score $s_{n,m}$ is modeled as follows. Assume that the $n^{th}$ participant exhibits an unknown \emph{preference weight} $\boldsymbol{\beta}_n = \{\beta_{n, 1}, \cdots, \beta_{n, L}\}$ over $L$ group fairness notions. In other words, $\beta_{n, l} \in [0, 1]$ and $\displaystyle \sum_{l = 1}^L \beta_{n,l} = 1$, for all $n, l$. Let $\phi_{\ell}(\boldsymbol{d}_m)$ denote the evaluation of ARP from the perspective of $\ell^{th}$ fairness notion. For the sake of brevity, the computation of group fairness notions is discussed in detail in Appendix \ref{App: Group Fairness Notions}. Let the $n^{th}$ participant aggregate the $L$ fairness evaluations of ARP as
\begin{equation}
\psi_{n,m}(\boldsymbol{\beta}_n) = \sum_{l = 1}^L \beta_{n, l} \cdot \phi_l(\boldsymbol{d}_m).
\label{Eqn: participant aggregate fairness evaluation}
\end{equation}
Since any fairness evaluation $\phi_l(\boldsymbol{d}_m)$ lies between $-1$ and $1$, the aggregated fairness evaluation $\psi_{n,m}(\boldsymbol{\beta}_n) \in [-1, 1]$. Consequently, if $\psi_{n, m}(\boldsymbol{\beta}_n) = 0$, the $n^{th}$ participant deems the ARP as a fair system. On the contrary, if $\psi_{n,m}(\boldsymbol{\beta}_n) = 1 \text{ or } -1$, the $n^{th}$ participant will deem the ARP system as an unfair one. However, the $n^{th}$ participant encodes their aggregated fairness evaluation $\psi_{n,m}(\boldsymbol{\beta}_n)$ using Likert scale and reports a fairness feedback score $s_{n,m} \in \{1, \cdots, 7\}$.

For the sake of simplicity, assume that the Likert encoding is accomplished by dividing the interval $[-1,1]$ into 14 equal partitions, each with width $\delta = 1/7$. The boundaries of these partitions are therefore given as $b_i = -1 + i \cdot \delta$ for all $i = 0, 1, 2, \cdots, 14$. Let $\mathbb{R}_i$ denote the union of two partitions corresponding to the interval $[b_{i-1}, b_i]$ and $[b_{14-i}, b_{14-i+1}]$, for all $i = 1, \cdots, 14$.

In practice, participants often compute a noisy fairness evaluation, due to the ambiguity in their preferences towards diverse fairness notions. This ambiguity in the preferences across fairness notions is modeled as follows. Let the true intrinsic fairness evaluation $\psi$ follow a logit-Normal distribution $F ( \cdot | \mu, \sigma )$, where the mean and variance of logit variable $\texttt{Logit}(\psi) = \log \frac{\psi}{1 - \psi}$ are given by $\mu = \psi_{n,m}(\boldsymbol{\beta}_n)$ and some known constant $\sigma^2$ respectively. Then, the $n^{th}$ participant experiences a utility $u_{n,i}$ as the probability of the true intrinsic fairness evaluation $\psi$ to lie in a specific region $\mathbb{R}_i$. In other words, the utility is formally given by
\begin{equation}
u_{n,m,i}(\boldsymbol{d}_m) = V_i \Big( \psi_{n,m}(\boldsymbol{d}_m) \Big) + V_{14 - i + 1} \Big(\psi_{n,m}(\boldsymbol{d}_m) \Big),
\label{Eqn: Utility}
\end{equation}
where 
\begin{equation}
\begin{array}{l}
V_i \Big( \psi_{n,m}(\boldsymbol{d}_m) \Big) = \displaystyle  F \left( \frac{1-b_i}{2}; \psi_{n,m} (\boldsymbol{d}_m) , \sigma \right) 
\\[2ex]
\qquad \qquad \qquad \qquad \displaystyle - \ F \left( \frac{1-b_{i-1}}{2}; \psi_{n,m} (\boldsymbol{d}_m) , \sigma \right)
\\[3ex]
\qquad \qquad \qquad \ \ = \displaystyle \int_{b_{i-1}}^{b_{i}} f(z; \psi_{n,m}(\boldsymbol{d}_m), \sigma) dz
\end{array}
\label{Eqn: Prob that psi in [b{i-1}, bi]}
\end{equation}
is the probability that the true intrinsic fairness evaluation $\psi$ lies in the interval $[b_{i-1}, b_i]$, where $f(\cdot; \mu, \sigma)$ is the logit-normal density function with parameters $\mu$ and $\sigma$. Then, the fairness feedback score $s_{n,m}$ is modeled as the logit probability
\begin{equation}
\tilde{s}_{n,m} = \displaystyle \frac{1}{\Delta_{n,m}} \cdot \Big\{  e^{\lambda\cdot u_{n, m, 1}}, \cdots,  e^{\lambda\cdot u_{n, m, 7}} \Big\},
\label{Eqn: Logit model}
\end{equation}
where $\Delta_{n,m} = \displaystyle \sum_{j = 1}^7 e^{\lambda\cdot u_{n, m, j}}$ is the normalizing factor, and $\lambda$ is the temperature parameter that captures the participant's sensitivity to the utilities.

\begin{figure}[!t]
\centering
\includegraphics[width = 0.45\textwidth]{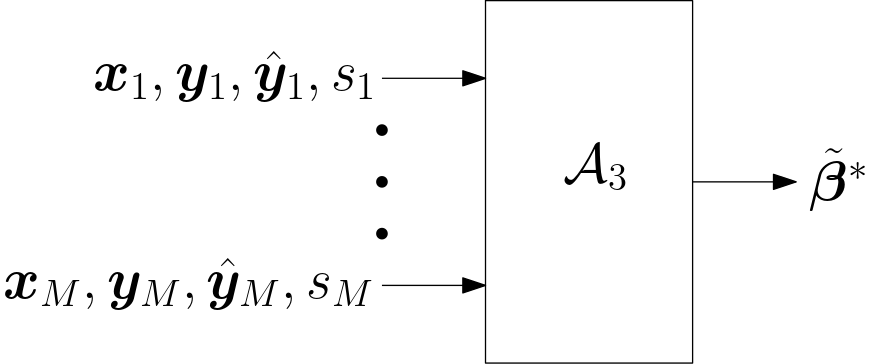}
\caption{Social Aggregation of Fairness Feedback}
\label{Fig: Method 2}
\end{figure}

\subsection{Proposed Algorithm}

The goal of this approach is to develop a social preference weight $\boldsymbol{\beta}^*$ that minimizes the average feedback regret $\mathcal{L}_F(\boldsymbol{\beta})$, which is given by
\begin{equation}
\mathcal{L}_F(\boldsymbol{\beta}) \triangleq \frac{1}{M} \sum_{m=1}^M \left( \frac{1}{N} \sum_{n = 1}^N \lVert s_{n, m} - \tilde{s}^*_m ( \boldsymbol{\beta} ) \rVert_2^2 \right),
\label{Eqn: Feedback Regret}
\end{equation}
where $\tilde{s}^*_m ( \boldsymbol{\beta} )$ represents the social fairness evaluation which follows the same definition in Equation \eqref{Eqn: Logit model}, but without having the participant index $n$. For the same reason, the participant index $n$ does not appear in Equations \eqref{Eqn: participant aggregate fairness evaluation}, \eqref{Eqn: Utility}, and \eqref{Eqn: Prob that psi in [b{i-1}, bi]} as well, for the computation of social fairness evaluation $\tilde{s}^*_m ( \boldsymbol{\beta} )$.

\begin{algorithm2e}[!t]
\caption{SAFF}
\label{Alg: SAFF}
\DontPrintSemicolon
\KwIn{$\mathbf{x}_{1:M}, \mathbf{y}_{1:M},\hat{\mathbf{y}}_{1:M}, \boldsymbol{s}_1, \ldots, \boldsymbol{s}_N, \delta$}
\KwOut{Learned social preference $\tilde{\boldsymbol{\beta}}^*$}
\BlankLine
Initialize $\boldsymbol{\beta}^{(0)}$ with a random $L$-dim. weight\;
\BlankLine
\For{$e = 1$ \KwTo num\_epochs}
{
\For{$m = 1$ \KwTo $M$}
{ 
$\boldsymbol{\phi}_m \leftarrow \texttt{FairnessScores}(\mathbf{x}_m, \mathbf{y}_m,\hat{\mathbf{y}}_m)$\; 
$\tilde{s}^*_m \leftarrow \texttt{EstimateFeedback}(\boldsymbol{\beta}^{(e)}, \boldsymbol{\phi}_m)$\;
\BlankLine
}
$\nabla \mathcal{L}_F(\boldsymbol{\beta}) \leftarrow \texttt{SRG}(s_{1, m}, \ldots, s_{N, m}, \tilde{s}^*_m, \boldsymbol{\phi}_m, \boldsymbol{\beta}^{(e)})$\;
$\boldsymbol{\beta}^{(e+1)} \leftarrow \mathbb{P} \left[ \boldsymbol{\beta}^{(e)} - \delta \cdot \nabla \mathcal{L}_F(\boldsymbol{\beta}) \right]$\;
}
\end{algorithm2e}

The social preference weight $\boldsymbol{\beta}^*$ can be learned using \emph{Social Aggregation of Fairness Feedback} (SAFF) algorithm as shown in Algorithm \ref{Alg: SAFF}, which is developed using projected gradient descent. The projection operator $\mathbb{P}$ ensures that $\boldsymbol{\beta}^*$ is a valid preference weight vector that has entries between 0 to 1 and sums to 1. The regret gradient $\nabla \mathcal{L}_F$ with respect to the model parameters $\boldsymbol{\beta}$ is computed using the well-known \emph{backpropagation} algorithm, as shown below:
\begin{subequations}
\begin{align}
\displaystyle \nabla_{\boldsymbol{\beta}} \mathcal{L}_F & = \displaystyle ( \nabla_{\tilde{s}^*} \mathcal{L}_F )^T \cdot \nabla_{\boldsymbol{\beta}} \tilde{s}^* \label{Eqn: regret grad beta}
\\[1ex]
\displaystyle \nabla_{\boldsymbol{\beta}} \tilde{s}^* & = \displaystyle (\nabla_{\boldsymbol{u}} \tilde{s}^* )^T \cdot \nabla_{\boldsymbol{\beta}} u \label{Eqn: s-tilde* grad beta}
\\[1ex]
\displaystyle \nabla_{\boldsymbol{\beta}} u & = \displaystyle (\nabla_{\boldsymbol{\psi}} u)^T \cdot \nabla_{\boldsymbol{\beta}} \boldsymbol{\psi} \label{Eqn: u grad beta}
\end{align}
\end{subequations}
where the gradient $\nabla_{\boldsymbol{q}} \boldsymbol{p}$ is a $P \times Q$ matrix, where $\boldsymbol{p}$ is a $P \times 1$ vector, and $\boldsymbol{q}$ is a $Q \times 1$ vector, for any general $\boldsymbol{p}$ and $\boldsymbol{q}$. Note that the gradients $\nabla_{\tilde{s}^*} \mathcal{L}_F$, $\nabla_{\boldsymbol{u}} \tilde{s}^*$, $\nabla_{\boldsymbol{\psi}} u$ and $\nabla_{\boldsymbol{\beta}} \boldsymbol{\psi}$ in Equations \eqref{Eqn: regret grad beta}, \eqref{Eqn: s-tilde* grad beta} and \eqref{Eqn: u grad beta} can be respectively computed as
\begin{equation}
\begin{array}{rl}
\nabla_{\tilde{s}^*} \mathcal{L}_F & = 
\displaystyle 2 \left[ \frac{1}{M} \sum_{m=1}^M \tilde{s}^*_m ( \boldsymbol{\beta} ) - \frac{1}{M N} \sum_{m=1}^M \sum_{n = 1}^N s_{n, m} \right],
\end{array}
\label{Eqn: regret grad s-tilde* - final}
\end{equation}
$\nabla_{\boldsymbol{u}_{m}} \tilde{s}_{m}^*$ is a $7 \times 7$ matrix, where the $(i,k)^{th}$ entry $\eta_{i,k}$ is given by 
\begin{equation}
\begin{array}{rl}
\eta_{i,k} 
& = 
\begin{cases}
\displaystyle \frac{\lambda}{\Delta_m^2} \cdot e^{\lambda u_{m,i}} \cdot \sum_{j \neq i} e^{\lambda u_{m,j}}, & \text{if } i = k,
 \\[2ex]
\displaystyle - \frac{\lambda}{\Delta_m^2} \cdot e^{\lambda u_{m,i}} \cdot e^{\lambda u_{m,k}}, & \text{otherwise},
\end{cases}
\end{array}
\label{Eqn: s-tilde* grad u - final}
\end{equation}
with $\Delta_m = \displaystyle \sum_{j = 1}^7 e^{\lambda\cdot u_{m, j}}$ being the normalizing factor,  
\begin{equation}
\begin{array}{l}
\nabla_{\psi_m} u_{m,i} = 
\displaystyle \frac{1}{\sigma^2} \left[ \frac{\sigma}{\sqrt{2\pi}} \exp \left\{ - \frac{(z_{i-1} - \psi_m)^2}{2 \sigma^2} \right\} \right.
\\[3ex]
\quad \displaystyle - \frac{\sigma}{\sqrt{2\pi}} \exp \left\{ - \frac{(z_i - \psi_m)^2}{2 \sigma^2} \right\} + \displaystyle \frac{\psi_m}{2} \erf \left(\frac{z_i - \psi_m}{\sigma \sqrt{2}} \right)
\\[3ex]
\quad  \qquad  - \displaystyle \frac{\psi_m}{2} \erf \left( \frac{z_{i-1} - \psi_m}{\sigma \sqrt{2}} \right) - \psi_m u_{m,i} 
\\[3ex]
\qquad \qquad \left. \displaystyle + \frac{\sigma}{\sqrt{2\pi}} \exp \left\{ - \frac{(z_{14-i} - \psi_m)^2}{2 \sigma^2} \right\} \right.
\\[3ex]
\qquad  \qquad \quad  \displaystyle - \frac{\sigma}{\sqrt{2\pi}} \exp \left\{ - \frac{(z_{14-i+1} - \psi_m)^2}{2 \sigma^2} \right\}
\\[3ex]
\qquad \qquad \qquad+ \displaystyle \frac{\psi_m}{2} \erf \left(\frac{z_{14-i+1} - \psi_m}{\sigma \sqrt{2}} \right)
\\[3ex]
\qquad \qquad \qquad \quad  \displaystyle \left. - \frac{\psi_m}{2} \erf \left( \frac{z_{14-i} - \psi_m}{\sigma \sqrt{2}} \right) \right],
\end{array}
\label{Eqn: u grad psi - final}
\end{equation}
where $z_i = \texttt{Logit}(b_i)$, and
\begin{equation}
\begin{array}{lcl}
\nabla_{\boldsymbol{\beta}} \psi_m = \boldsymbol{\phi}(\boldsymbol{d}_m).
\end{array}
\label{Eqn: psi grad beta - final}
\end{equation}

The method of computing the gradient of social regret is called \emph{Social Regret Gradient} (SRG), which is formally presented in Algorithm \ref{Alg: SRG}.

\begin{algorithm2e}[!t]
\caption{SRG}
\label{Alg: SRG}
\DontPrintSemicolon
\KwIn{$\boldsymbol{s}_1, \ldots, \boldsymbol{s}_N, \tilde{s}^*, \boldsymbol{\phi}, \boldsymbol{\beta}$}
\KwOut{Feedback Regret Gradient $\nabla \mathcal{L}_F(\boldsymbol{\beta})$}
\BlankLine
Compute $\nabla_{\boldsymbol{\beta}} \psi_m$ using the Equation \eqref{Eqn: psi grad beta - final}\;
Compute $\nabla_{\psi_m} u_{m,i}$ using the Equation \eqref{Eqn: u grad psi - final}\;
Compute $\nabla_{\boldsymbol{u}_{m}} \tilde{s}_{m}^*$ using the Equation \eqref{Eqn: s-tilde* grad u - final}\;
Compute $\nabla_{\tilde{s}^*} \mathcal{L}_F$ using the Equation \eqref{Eqn: regret grad s-tilde* - final}
\end{algorithm2e}

\section{Evaluation Methodology \label{sec: Evaluation}}
The proposed algorithm SAFF is employed on both simulated data as well as survey responses. This paper considers $L = 6$ group fairness notions (see Table \ref{Tab: Group fairness notions}) to evaluate the Acceptance Rate Predictor (ARP) with respect to the sensitive attributes \emph{race} =\{Black, All Other Races\}, \emph{gender} = \{Male, Female\}, and \emph{age} = \{$<$50, $>$50\}. In addition, the privileged and underprivileged groups are defined as $\mathcal{X}_m$ = \{Other, Male, $<$50\} and $\mathcal{X}_{m'}$ = \{Black, Female, $>$50\}, respectively. The predicted probability of kindey acceptance from the ARP is discretized into binary, where the probability $\geq 0.5$ indicates acceptance ($\hat{y} = 1$), and probability $<0.5$ indicates rejection ($\hat{y} = 0$). The computation of various group fairness scores is elaborated in Appendix \ref{App: Group Fairness Notions}. 

\begin{figure*}[!t]
\floatconts
{Fig: Beta, Feedback Regret}
{\vspace{-4ex}\caption{Convergence of Feedback Regret across Different Data-Tuple Sizes}}
{%
\subfigure[Age Attribute]{%
\label{Fig: Feedback Regret - Age}
\includegraphics[width=\textwidth]{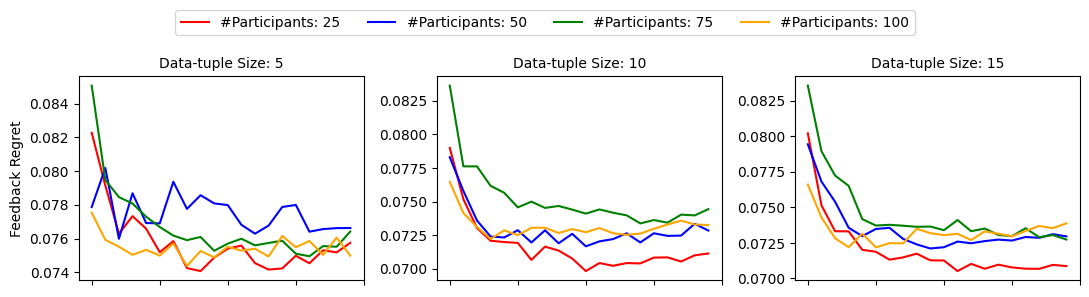}
}\hfill 
\subfigure[Race Attribute]{%
\label{Fig: Feedback Regret - Race}
\includegraphics[width=\textwidth]{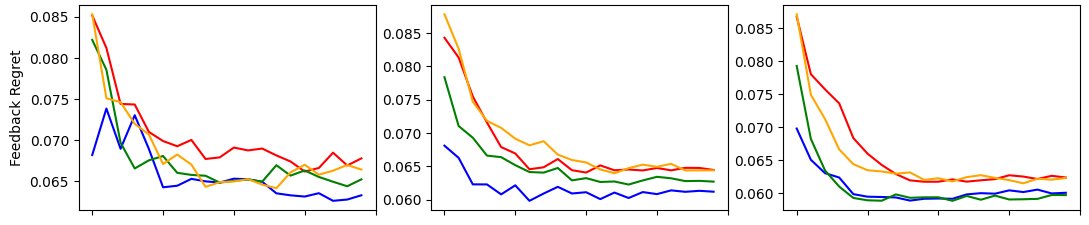}
}\hfill 
\subfigure[Gender Attribute]{%
\label{Fig: Feedback Regret - Gender}
\includegraphics[width=\textwidth]{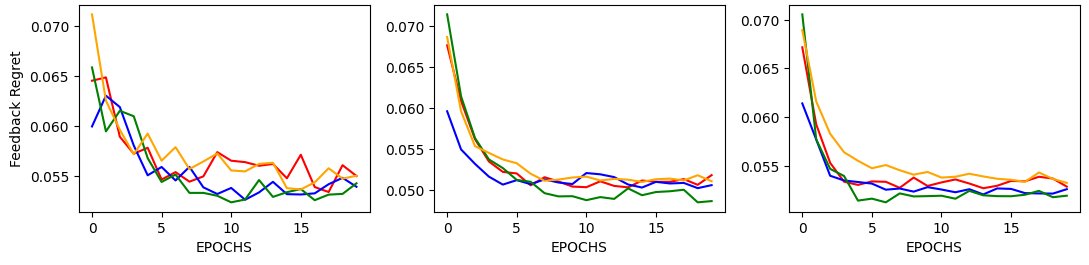}
}
}
\end{figure*}

\subsection{Evaluation on Simulated Data}

For simulation experiments, the true preferences of the non-expert participants $\boldsymbol{\beta}_1, \boldsymbol{\beta}_2, \ldots, \boldsymbol{\beta}_N$ are constructed by randomly assigning preference values for all $L = 6$ fairness notions based on uniform distribution. Similarly, the estimated social preference is also initialized with
random values based on uniform distribution. The estimated social preference $\boldsymbol{\beta}^*$ is updated over $M = \{5, 10, 15\}$ data tuples each containing $K = 10$ donor-recipient pairs. The results are averaged across 100 iterations for all $N = \{25, 50, 75, 100\}$ non-expert participants. The learning rate is declared as $\delta = 0.1$ and the number of epochs as 20.


\subsection{Evaluation on ARP Survey}
Unlike simulation experiment, the true preferences of the participants are unknown in the survey experiment. The estimated social preference $\boldsymbol{\beta}^{(0)}$ is initialized randomly based on uniform distribution. Note that the participants rate the fairness of ARP on a Likert scale of 1 to 7, $\boldsymbol{s}_n \in \{1, 2, \cdots, 7\}$. The estimated social preference $\boldsymbol{\beta}^{(0)}$ is updated over $M = 10$ data-tuples each containing $K = 10$ donor-recipient pairs presented to $N = 75$ participants. 

\section{Results and Discussion \label{sec: Results}}

\begin{figure*}[!t]
\centering
\includegraphics[width = \textwidth]{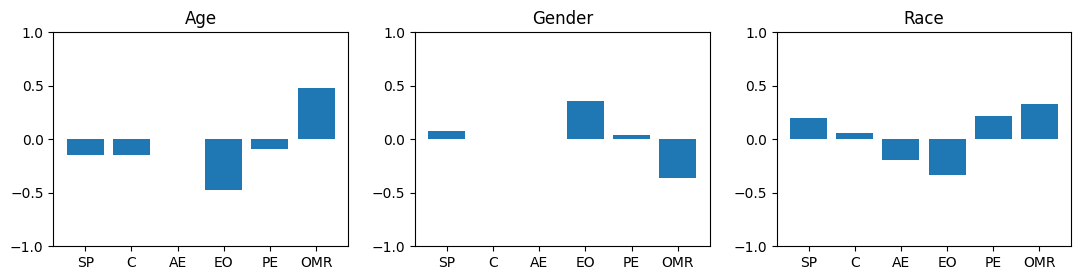}
\caption{Group Fairness Evaluations of the ARP across Different Sensitive Attributes.}
\label{Fig: ARP Fairness}
\end{figure*}

\begin{table*}[!t]
\centering
\begin{tabular}{c c c c c c c}
\toprule
\textbf{Sensitive Attribute} & \multicolumn{6}{c}{\textbf{Social Fairness Preference}}
\\ \midrule\midrule
& \textbf{SP} & \textbf{C} & \textbf{AE} & \textbf{EO} & \textbf{PE} & \textbf{OMR}
\\ \midrule
Age & 0.15 & 0 & \textbf{0.45} & 0.007 & 0.37 & 0.01
\\
Gender & 0.19 & 0.02 & \textbf{0.48} & 0 & 0.24 & 0.06
\\
Race & 0.28 & 0.10 & \textbf{0.38} & 0 & 0.19 & 0.03
\\ \bottomrule
\end{tabular}
\caption{Social Fairness Preferences of the Recruited Participants over $L = 6$ Group Fairness Notions}
\label{Tab: SAFF Survey Prefs}
\end{table*}

\subsection{Simulation Results}
Figure \ref{Fig: Beta, Feedback Regret} illustrates feedback regret for varying numbers of participants, $N = \{25, 50, 75, 100\}$, with each receiving $M = \{5, 10, 15\}$ data-tuples. Figure \ref{Fig: Feedback Regret - Age} demonstrates the social feedback regret with respect to the age attribute computed using the participants' responses to the question Q2 (refer Figure \ref{Fig: Survey Questions}). Similarly, Figure \ref{Fig: Feedback Regret - Race} depicts the social feedback regret with respect to the race computed using the responses received from question Q3. On the other hand, Figure \ref{Fig: Feedback Regret - Gender} shows the convergence of social feedback regret with respect to the gender computed using the responses from question Q4.

Note that the preference regret converges with increasing number of epochs for any sensitive attribute and any combination of data tuple size and the number of participants. However, the increase in the number of participants and/or data tuple size has little improvement on social feedback regret.


\textbf{Initialization:} The proposed algorithm converges quite well, as demonstrated in Figure \ref{Fig: Beta, Feedback Regret}, when the preference weights in the proposed model are initialized as random weight vectors. However, the same approach does not exhibit the desired convergence when the social preferences are initialized to equal preference, i.e. $\beta_l = 1/6$ for all $l = 1, \cdots, 6$.


\subsection{Survey Results}
Table \ref{Tab: SAFF Survey Prefs} shows the estimated social preferences of the recruited participants over $L = 6$ group fairness notions in the Prolific survey experiment. Note that \emph{accuracy equality} (AE) is the preferred group fairness notion across all three sensitive attributes. Note that the ARP is perceived to exhibit less bias in terms of accuracy equality across all three sensitive attributes (as shown in the Figure \ref{Fig: ARP Fairness}). In the case of age and gender, \emph{predictive equality} (PE) has the second highest preference over the six group fairness notions. Even from the perspective of PE, the ARP exhibits little/no bias wit respect to all the three sensitive attributes. On the contrary, although the ARP is perceived to have no bias in terms of \emph{calibration}, the social fairness preference is close to zero with respect to both age and gender.

At the same time, the ARP seems unfair in terms of \emph{equal opportunity} (EO) with evaluations ranging to $-0.5$ with respect to age, and $0.46$ with respect to gender (as depicted in Figure \ref{Fig: ARP Fairness}). However, EO is the least preferred fairness notion, with almost negligible preference weight for all the three sensitive attributes, as shown in the Table \ref{Tab: SAFF Survey Prefs}. Similar observations can be made with \emph{overall misclassification rate} (OMR) as well. Although the ARP is unfair in terms of OMR, the non-expert participants clearly do not prefer OMR. Therefore, group fairness notions such as C, EO and OMR have little role in public's fairness evaluation regarding the U.S. kidney placement.

In summary, accuracy equality and predictive equality can be deemed as critical group fairness notions from the public stakeholders' viewpoint. Furthermore, as a follow-up to the above claim, it is also natural to conclude that the non-expert participants' perceive ARP as a \textbf{reasonably} fair system when deployed in the kidney placement pipeline. 

\section*{Acknowledgments}
This work is supported by the National Science Foundation (Award \#2222801). We express our gratitude to the authors \cite{ashiku2022identifying} for granting access to their proposed machine learning model. Additionally, we extend our appreciation to the United Network of Organ Sharing (UNOS) for support this project and for providing the kidney matching datasets.

\bibliography{sample-bibliography}

\appendix

\begin{figure*}[!t]
\centering
\includegraphics[width=\textwidth]{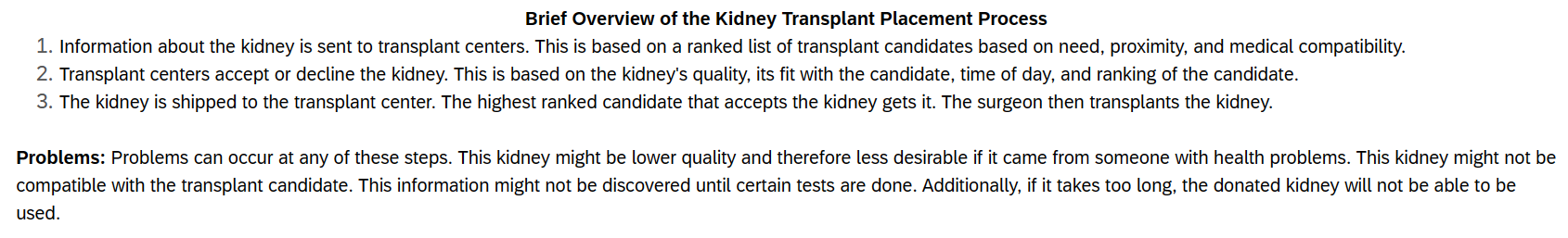}
\includegraphics[width=\textwidth]{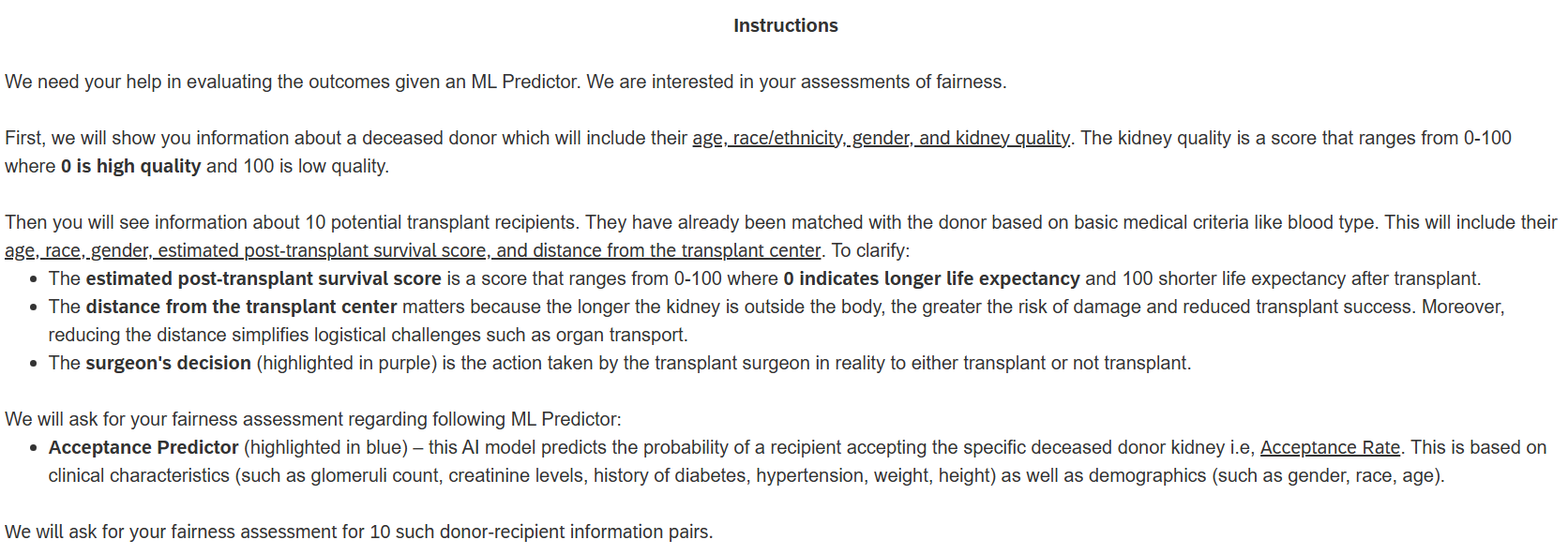}
\caption{Kidney Placement Overview and the Survey Instructions Presented to the Participants.}
\label{Fig: Survey Instructions}
\end{figure*}

\section{Kidney Placement in the United States \label{Sec: Kidney Placement}}
The term kidney placement refers to the process of procuring kidneys and identifying potential recipients for transplant surgery based on several donor/recipient characteristics, as well as location proximity. In the United States, organ procurement and transplantation are led by the United Network of Organ Sharing (UNOS), where donors in the Organ Procurement Organizations (OPOs) are matched with patients waiting for organs in the Transplant Centers (TXCs). The OPOs are responsible for procuring the organs, evaluating them for quality using Kidney Donor Profile Index (KDPI) score, and maintaining a donor registry. The KDPI score, ranging from 0 to 100, is computed using donor characteristics such as donor's age, height, race, and history of hypertension, where 0 indicates high quality and 100 indicates low quality. On the other hand, the TXCs are responsible for evaluating recipients on the waiting list using Estimated Post Transplant Survival (EPTS) score and performing transplant surgery. The EPTS score, also ranging from 0 to 100, is computed using patient attributes such as patient's age, years on dialysis, and diabetes status, where 0 implies longer life expectancy and 100 implies shorter life expectancy. Once a deceased donor kidney is identified as suitable, it will be matched with the candidates in the waiting list based on scores computed from KDPI and EPTS \citep{friedewald2013kidney}. Thereafter, the potential recipients for a specific deceased donor kidney are ranked based on geographic location and medical urgency. As of now, a single deceased donor kidney can be matched with thousands of potential recipients and at most two of them will undergo kidney transplantation.


\section{Survey Information \label{App: Complete survey}}
First, the recruited participants are presented with a brief overview of the kidney placement process in the United States which includes information regarding the transplant centers, kidney offers, identifying potential recipient, and transportation of the donor kidney. In the next page, instructions regarding the survey experiment is detailed. Specifically, this page explains how the data-tuple is represented, different donor-recipient attributes involving in a data-tuple, and what is expected from the participants (as shown in Figure \ref{Fig: Survey Instructions}).


\begin{table*}[!t]
\caption{Diverse Group Fairness Notions}
\centering
\resizebox{2\columnwidth}{!}{%
\begin{tabular}{c l l}
\toprule
Index ($l$) & \textbf{Group Fairness Notion ($f$)} & \textbf{Groupwise Rate $\phi_f(m)$}
\\\midrule \midrule
1 & Statistical Parity (SP) \citep{dwork2012fairness} & $\phi_{SP}(m) = \mathbb{P}(\hat{y} = 1 \ | \ x \in \mathcal{X}_m)$
\\[1ex]
2 & Calibration (C) \citep{chouldechova2017fair} & $\phi_{C}(m) = \mathbb{P}(y = 1 \ | \ \hat{y} = 1, x \in \mathcal{X}_m)$
\\[1ex]
3 & Accuracy Equality (AE) \citep{berk2018riskassess} & $\phi_{AE}(m) = \mathbb{P}(\hat{y} = y \ | \ x \in \mathcal{X}_m)$
\\[1ex]
4 & Equal Opportunity (EO) \citep{MoritzOpportunities} & $\phi_{EO}(m) = \mathbb{P}(\hat{y} = 1 \ | \ y = 1, x \in \mathcal{X}_m)$
\\[1ex]
5 & Predictive Equality (PE) \citep{corbett2017algorithmic} & $\phi_{PE}(m) = \mathbb{P}(\hat{y} = 1 \ | \ y = 0, x \in \mathcal{X}_m)$
\\[1ex]
6 & Overall Misclassification Rate (OMR) \citep{rouzot2022learning} & $\phi_{OMR}(m) = \mathbb{P}(\hat{y} = 0 \ | \ y = 1, x \in \mathcal{X}_m)$
\\\bottomrule
\end{tabular}
}
\label{Tab: Group fairness notions}
\end{table*}

\section{Group Fairness Notions \label{App: Group Fairness Notions}}
Over the past decade, several group fairness notions have been proposed to measure the biases in a given system. Such fairness notions seek for parity of some statistical measure (e.g. true positive rate, predictive parity value) be equal across all the sensitive attributes (e.g. race) present in the data. Specifically, group fairness notions measure the difference in a specific statistical measure between protected (e.g. Caucasians) and unprotected (e.g. African-Americans) groups of a sensitive attribute. Different versions of group-conditional metrics led to different statistical definitions of fairness \cite{caton2020fairness,chouldechova2018frontiers,mehrabi2021survey,pessach2020algorithmic}. Let $y \in \mathcal{Y}$ as the true label and $\hat{y} = g(x) \in \mathcal{Y}$ as the predicted label given by the ML-based system for some input $x \in \mathcal{X}$. Furthermore, let $\mathcal{X}_m, \mathcal{X}_{m'} \in \mathcal{X}$ denote the protected and unprotected sensitive groups respectively. The \emph{unfairness} within the acceptance predictor can be evaluated based on several group fairness notions which can be generalized as 
\begin{equation}
\phi_f \triangleq \ \phi_f(m) - \phi_f(m'),
\end{equation}
for any $\mathcal{X}_m, \mathcal{X}_{m'}$, and $\phi_f(m)$ denotes the groupwise rate with respect to the group $\mathcal{X}_m$. Various groupwise rates studied in the literature are listed in Table \ref{Tab: Group fairness notions}. 

\end{document}